\documentclass[letterpaper, 10 pt, journal, twoside]{IEEEtran}

\usepackage{amsmath} 
\usepackage{amssymb}  
\usepackage{graphicx}
\graphicspath{{figures/}}
\usepackage[font=small,skip=2pt]{caption}
\usepackage{subcaption}
\usepackage{bbold} 
\usepackage{tabularx}
\usepackage{svg}
\usepackage{comment}
\usepackage{xcolor}
\usepackage{import}
\usepackage{multirow}
\usepackage{microtype}
\usepackage{xifthen}
\usepackage{dblfloatfix}
\usepackage{soul}
\usepackage{mathrsfs}
\usepackage{xifthen}
\usepackage{pdfpages}
\usepackage{transparent}

\usepackage{algorithm}
\usepackage{algorithmicx}
\usepackage{algpseudocode}
\floatname{algorithm}{Procedure}

\usepackage{float}

\usepackage[
giveninits=true,
maxbibnames=99,
backend=biber,
style=ieee,
sorting=none
]{biblatex}
\newcommand{\methodname}{GELATO}

\title{ \methodname: Multi-Instruction Trajectory Reshaping via Geometry-Aware Multiagent-based Orchestration \vspace{-5pt}}

\author{Junhui Huang$^{1,2}$, Yuhe Gong$^{2}$, Changsheng Li$^{1}$, Xingguang Duan$^{1}$, Luis Figueredo$^{2,3}$%
\thanks{$^{1}$ Junhui Huang Changsheng Li and Xingguang Duan are with the Beijing Institute of Technology (BIT). $^{2}$The authors are with the School of Computer Science, University of Nottingham, UK. $^{3}$ Figueredo is also an Associated Fellow at the Munich Institute of Robotics and Machine Intelligence (MIRMI), at the Technical University of Munich (TUM). 
This work was partially funded by the Lighthouse Initiative Geriatronics by StMWi Bayern (Project X, grant no. IUK-1807-0007// IUK582/001). Email:\texttt{Junhui.Huang}@nottingham.ac.uk, \texttt{lics@}bit.edu.cn, \texttt{yuhe.gong}@nottingham.ac.uk,  \texttt{duanstar}@bit.edu.cn,  \texttt{figueredo}@ieee.org  
} 

}

\colorlet{jh}{magenta}
\colorlet{lf}{orange}
\colorlet{ab}{red}

\newcommand{\removetext}[1]{}
\newcommand{\responseMarks}[1]{} 

\renewcommand{\baselinestretch}{0.93}

\makeatletter
\renewcommand{\section}{\@startsection{section}{1}{\z@}{1.0ex plus 0.8ex minus 0.5ex}%
	{0.4ex plus 0.5ex minus 0ex}{\normalfont\normalsize\centering\scshape}}%
\renewcommand{\subsection}{\@startsection{subsection}{2}{\z@}{0.5ex plus 0.5ex minus 0.5ex}%
	{0.3ex plus 0.4ex minus 0ex}{\normalfont\normalsize\itshape}}%

\makeatother

\begin{document}

\maketitle


\begin{abstract}
%
%
We present \methodname---the first 
language-driven trajectory reshaping framework to embed geometric environment awareness and multi-agent feedback orchestration to support multi-instruction in human-robot interaction scenarios. 
%
Unlike prior learning-based methods,
our approach automatically registers scene objects as 6D geometric primitives via a VLM-assisted multi-view pipeline, and an LLM translates free-form multiple instructions into explicit, verifiable geometric constraints. 
These are integrated into a geometric-aware vector field optimization to adapt initial trajectories while preserving smoothness, feasibility, and clearance.  
%
We further introduce a multi-agent orchestration with observer-based refinement to  handle multi-instruction inputs and interactions among objectives---increasing success rate without retraining.  
Simulation and real-world experiments demonstrate our method achieves smoother, safer, and more interpretable trajectory modifications compared to state-of-the-art baselines.
%
%
\end{abstract}


\section{Introduction}


As robots move from structured cells to dynamic, human-centric spaces, intuitive and seamless interaction with people and their surroundings becomes paramount.
A critical aspect of this interaction is adapting movements in response to human preferences, concerns, and instructions. Such corrective, language-driven feedback is central to building robots that are robust, reliable, interactive, and adaptable---and it defines the problem addressed herein.

\begin{figure}[t!]
		\centering
        \includegraphics[width=0.98\columnwidth]{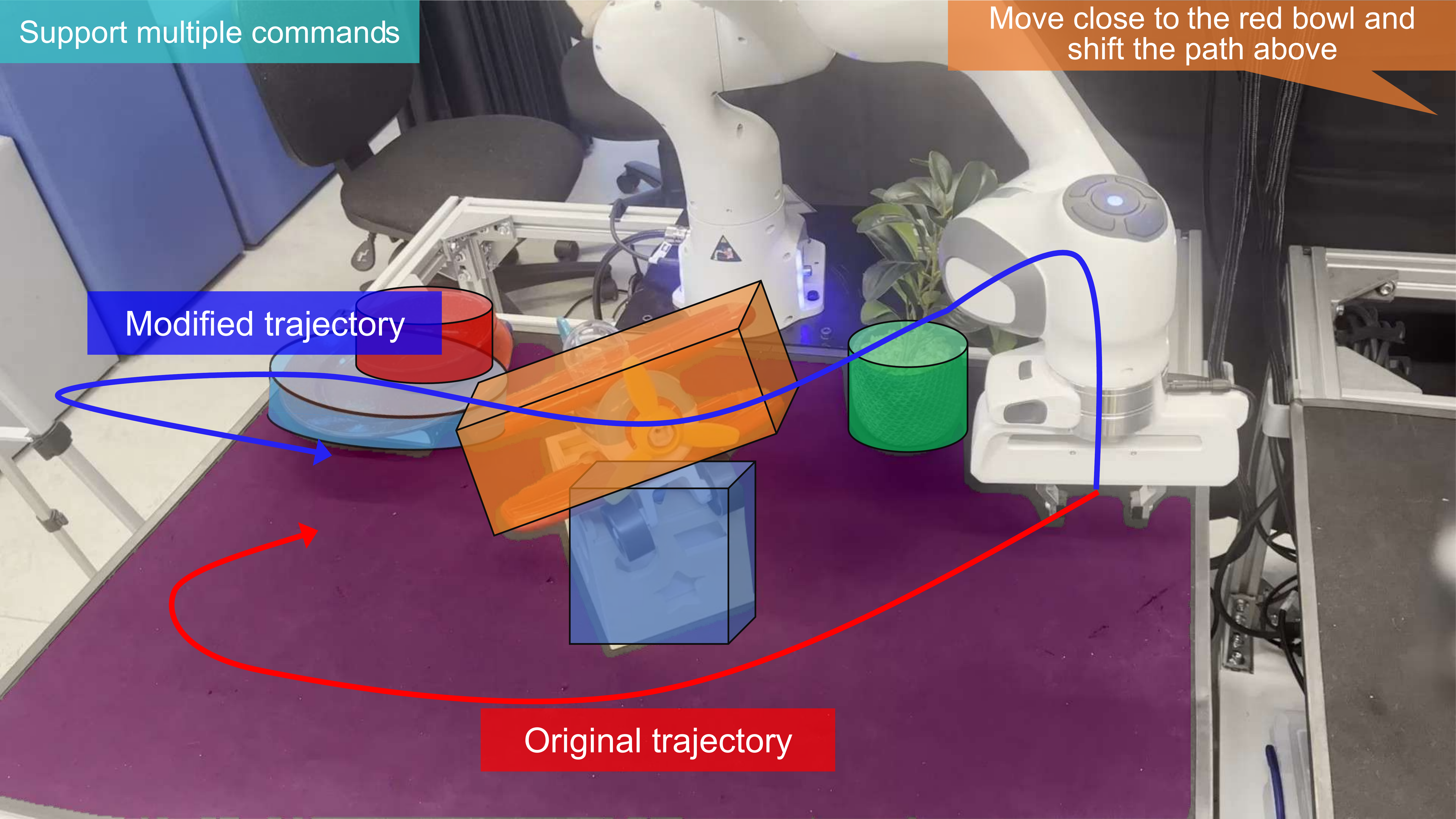}
        \caption{Multi-instructions trajectory adaptation based on natural language commands and 6D environment information.  \methodname~is the first language-guided trajectory adaptation framework to geometric-aware interactions (considering full object poses instead of keypoints)  
        and proper safe modulation.  The framework fuses object semantics with automatic geometric registration (planes, cylinders, cuboids) and safe modulation to satisfy multiple instructions. A multi-agent loop proposes alternative semantic-based strategies and an observer verifies constraint satisfaction to refine the final path.
}
		\label{geo_spot}
    \end{figure}

Recent progress in learning by demonstration \cite{ravichandar2020recent}, novel algorithms for imitation learning \cite{driess2023palm,kim2025openvla}, and the emergence of foundation models \cite{mousavian20196,ravi2024sam,gadre2022clip} applied in robotics  points to a promising path, enabling robots to perform increasingly complex tasks in human environments. 
However, despite these advances, most solutions are still deployed open-loop, with pre-programmed actions and little to no ability to adapt on the fly to contextual changes.
Robots need a more flexible framework that allows them 
%
to continuously refine their actions based on changing context, goals, and human inputs.  
Indeed, this kind of on-the-fly corrective feedback is one of the hallmarks of human behaviour
%
Consider a dinner-table scenario, where you may be asked to move an object closer to a bowl as it will be served there, while keeping the path above the table to make it shorter and more interpretable (see Fig.~\ref{geo_spot}).    
This work targets precisely this facet of human–robot interaction (HRI), i.e.,  efficient, reliable, interpretable, interactive language-driven adaptations for trajectory reshaping.

Our goal is to enable robots to understand, reason about, and respond on the fly to natural-language instructions, while also recognising and respecting the geometry and constraints of the surrounding scene, previous constraints, and natural HRI objectives such as moving smoothly and avoiding collisions.  
Meeting this request requires 
(i) understanding the instruction in open vocabulary; 
(ii) reasoning over the full 6D extent of objects and the scene---going beyond existing language-driven models that treat objects as single points (e.g., “move away from the table” presumes understanding the table is shaped as a plane/cuboid, not a single point);  
(iii) translating language into explicit, interpretable, and verifiable motion constraints tied to object semantics; and  
(iv) robustly resolving multiple, potentially interacting instructions.

These requirements, translated to zero-shot handling of free-form language, geometry-aware grounding, interpretability, and reliable execution under multi-objective commands, define our proposed framework \methodname~(GEometric-aware Language-guided and Agent-based Trajectory mOdulation). 

In brief, in our approach,  
%
a VLM-assisted registration firstly converts multi-view RGB-D into a primitive-based 6D scene, 
enabling closest-point reasoning and geometry awareness beyond single point-to-point schemes.  
Next, an LLM translates free-form instructions into explicit constraints connected to those primitives and to  
vector-field optimizers. These reshape the trajectory so that interpretable instructions and constraints are satisfied, while preserving length/curvature, 
object clearance, and improving smoothness.  
To tackle long-horizon or interacting instructions, 
\methodname~runs multiple agents based on different reasoning around the language instructions (e.g., assuming sequential or parallel actions, giving priority or equal weight). Finally, an observer verifies each instruction and constraint 
(e.g., distance, directional displacement, local speed) and, if any check fails, a refinement loop adjusts weights/ranges and retries—increasing success without retraining.

Our framework introduces the following contributions and novelties to language-driven robotics.
\begin{itemize}
    \item  \textit{VLM-geometric-aware scene registration} grounded into analytic primitives, targeting safe and efficient trajectories; 
    \item \textit{Interpretable language-driven grounding} for trajectory reshaping, enabling verifiable geometric and kinematic objectives tied to semantics and interactive vector-field optimisation---ensuring safety, clearance and smoothness.  
%
    %
\item  \textit{Multi-agent interactive feedback}, 
providing diverse reasoning coverage for complex/multi-instruction tasks  
together with an observer–refinement feedback-loop; 
%
%
\item 
Novel representative datasets, comprehensive validation through simulations, a user study, and real-robot trials. 
\end{itemize}

\section{Related Work }

\begin{figure*}[h!]
		\centering
		\includegraphics[width=18cm]{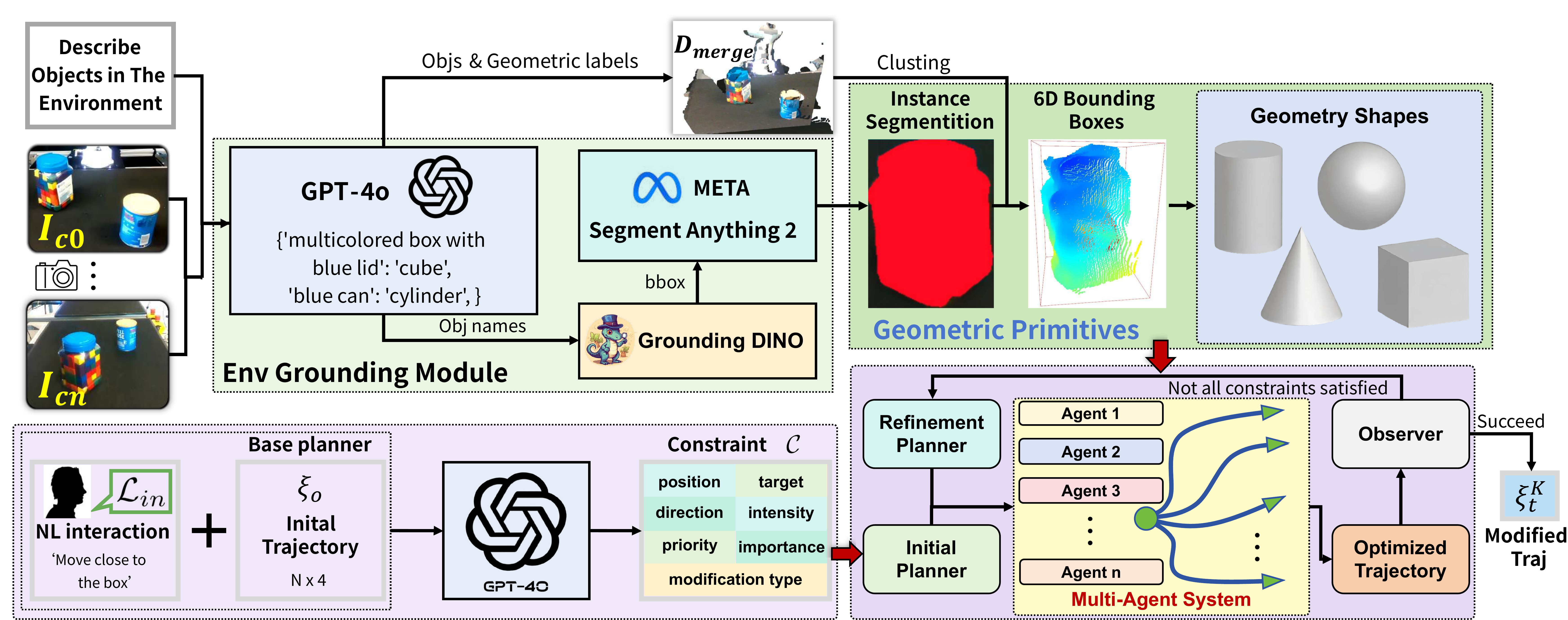}
		\caption{System Architecture. 
        (\textbf{Top}) 
        The environment-grounding module segments objects (Grounding-DINO, SAM), fuses multi-view RGB-D, and fits geometric primitives (6D OBB), producing a structured scene. 
        (\textbf{Bottom}) Translates language instructions into structured motion constraints using an LLM.
        (\textbf{Right}) A multi-agent system that integrates the outputs from both streams to reason over and interactively optimize an initial trajectory, producing a final path that fulfils the user's intent. The process iterates until all constraints are satisfied.
        %
        %
        }         
        \label{fig:model_arc}
\end{figure*}

\subsection{Natural Language Interfaces for Robotic Interaction}

A fundamental challenge in enabling intuitive human-robot interaction is grounding the rich semantics of natural language within the physical and geometric constraints of a human-centered environment. Early methods were limited, either by confining users to rigid command structures \cite{tellex2020robots} or employing complex probabilistic models that were difficult to scale \cite{arkin2020multimodal,walter2021language}. More recently, a dominant paradigm has emerged: using deep neural networks, including large-scale foundation models, to learn a direct, end-to-end mapping from language to robot actions \cite{fu2019language,hong2011recurrent}.

{However, a primary drawback of these approaches is their significant data requirement. To mitigate} this constraint, 
recent research,  \cite{bucker2022latte,bucker2022reshaping,stepputtis2020language}, has 
explored the use of 
large foundation models~\cite{devlin2018bert,gadre2022clip}, improving system generalization. These methods 
allowed training  
without specific annotated datasets, however, they still require extensive training---mostly replacing datasets with procedure training over LLMs. 
Furthermore, their fixed, post-training parameters limit their adaptability to new environments, and their reliance on a direct alignment between commands and objects often fails when interpreting complex user intent, leading to unpredictable results. This motivates a different paradigm to enhance the reliability of intent recognition, i.e., {using LLMs to reason about independent multiple preferences and constraints  
for trajectory modification---providing a more interpretable and controllable outcome which can be evaluated and adapted online. }

\subsection{Geometry for Robotics} 
Geometry has long played a pivotal role in robotics, 
encoding constraints, object modeling, and spatial reasoning through geometric predicates \cite{somani2017exact,chi2023geometric}.
%
Efficient convex decomposition techniques and model-based representations (e.g., 
\cite{montanari2017improving,funes2019novel,oleynikova2017voxblox}) are still largely deployed in modeling of complex environments \cite{oleynikova2017voxblox}. 
%
%
%
%
%
However, most methods, including iterative schemes like GJK \cite{montanari2017improving,montaut2024gjk++}---prioritize fast nearest-feature queries using abstract proxies (simplices, bounding volumes), yielding limited interpretability or generalization of the object information.  
Even modern approaches based on learning surrogates and voxelized signed distance fields \cite{oleynikova2017voxblox,curobo_icra23,staroverov2024dynamic}, which produce smooth, geometry-aware distances and gradients, still typically collapse to a scalar closest-point distance with purely local gradients without any semantics. As a result, the object’s shape, structure, and spatial configuration (and their rich semantics) inform the closest-point query but are largely discarded thereafter.

%


%

In contrast, a novel paradigm for encoding geometric-aware information based directly on geometric primitives have emerged~\cite{gong2025geopfinfusinggeometrypotential}. 
Therein, gradients and field-actions are shaped not only by the closests points, but also by the semantics of 
%
%
different geometric primitive distances---leading to interpretable and more efficient results. 
However, they 
neglect the geometry feature extraction---unrealistic assuming the geometry shape of the objects is fully known. In the context of feature extraction, RANSAC~\cite{derpanis2010overview} extracts primitives from raw point clouds, but it does not impose any size constraints. Point cloud-based primitive extraction methods~\cite{schnabel2007efficient,li2019supervised} can provide more detailed and refined geometric decomposition, however, they are sensitive to noise from depth sensors. SuperDec~\cite{fedele2025superdec} has also been shown to efficiently 
depict geometric distances---yet, through complex surfaces  which makes it difficult to explore primitives, limiting interpretability. 

Most importantly, 
none of such geometric-aware representations 
have been properly integrated with trajectory reshaping---despite its critical role in ensuring interpretability and safety. 
For instance, CLIP\cite{gadre2022clip} provides a single point with the pose of an object---fully neglecting the geometry information.  
An efficient trajectory modification must integrate such geometric information, ensuring robust and reliable trajectory-based feature attraction and obstacle modeling with visual environmental understanding.

\subsection{Large Language Model for Robotics}
Recent advancements in LLMs and VLMs have significantly enhanced robotic capabilities in spatial understanding and high-level task planning. A prominent line of research involves end-to-end Visual-Language-Action (VLA) models, such as PaLM-E \cite{driess2023palm} and openVLA \cite{kim2025openvla}, which directly map multimodal inputs to low-level actions through implicit latent representations. Other paradigms leverage LLMs as intermediate tools. For instance, Code-as-Policies \cite{liang2022code} translates tasks into interpretable code that orchestrates predefined motion primitives, while works like Voxposer \cite{huang2023voxposer}, RobotGPT \cite{jin2024robotgpt} and Rekep \cite{huang2024rekep} parse user intent into a sequence of API calls. For manipulation, systems such as VISO-Grasp \cite{shi2025viso} and ThinkGrasp \cite{qian2024thinkgrasp} merge reasoning of VLMs with specialised models like GraspNet \cite{mousavian20196} to perform complex grasping in cluttered environments.

A common thread in these approaches is a focus 
on multimodal  
trajectory generation, often abstracting away the fine-grained geometric features of objects. In contrast, our work addresses the specific challenge of modifying complex trajectories according to nuanced user preferences. We leverage the spatial reasoning of foundation models to create an automated visual registration pipeline, yielding a geometric representation for objects. Concurrently, an LLM interprets complex user commands to generate a diverse set of initial conditions that drive a multi-agent system. This allows our method to robustly adapt complex trajectories, ensuring the final path is precisely aligned with the user's intent.

\section{Approach}  

Our approach presents a comprehensive framework for modifying robot trajectories based on natural language commands, ensuring the final path is both compliant with user intent and kinematically feasible. To this aim, we devise a 
two-stage process: first,  building a detailed geometric-informed representation  
of the environment and the robot's trajectory therein, and second, refining the initial trajectory according to user instructions while accounting to the system representation. 

To achieve scene understanding, we introduce an automated pipeline that leverages a VLM, along with foundational vision models to register objects and their geometric properties.   
For trajectory modification, a separate LLM translates the user's command into a set of soft constraints. These constraints are then integrated into a geometric-aware vector field-based optimisation that iteratively adjusts the trajectory. This method balances external forces derived from the user's command and obstacle avoidance with internal forces that maintain the trajectory's smoothness and integrity. To robustly handle complex or conflicting instructions, we introduce a novel multi-agent system where different agents attempt to solve the constraints using varied strategies—such as parallel, sequential, or LLM-prioritised execution, thereby increasing the success rate for complex, multi-constraint tasks.




\subsection{Problem Definition}
Our core objective is to reshape a robot's trajectory using natural language commands. We decompose this challenge into two sequential sub-problems. 

\textbf{1) Geometric Scene Registration:} The first objective is to convert raw visual sensor input into a structured, robot-friendly geometric representation of the environment. Given a set of RGB-D images  $I_c = \{ I_{c0},I_{c1},\dots,I_{c|c|} \}$ from the same scene (stemming from two or more cameras after registration alignment), where $|c|$ is the number of cameras/images for each scene timestep, we seek to define  a function $g_{vlm}$ that grounds the scene into a set of $M$ geometric-primitive-oriented objects, $\mathcal{O} = \{O_1, \dots, O_M\}$. Each object $O_M \in \mathcal{O}$ is thus represented by its closest geometric primitive (e.g., cube, cylinder) defined by its merged/filtered pointcloud $D_{merge}$ and its own semantics  and its corresponding 6D pose (position and orientation). This mapping is defined as:
\begin{equation}
\mathcal{O} = g_{vlm}(I_{c},D_{merge}) .
\end{equation}

\textbf{2) Multi-Instruction Trajectory Reshaping:} The second objective is to adapt an initial trajectory according to a set of complex, and potentially conflicting, user instructions. The initial trajectory is defined as $\xi_o: [-1, 1] \rightarrow \mathbb{R}^{N \times 4}$, which is discretized into a sequence of $N$-waypoints, represented as $\xi_o = \{(x_i, y_i, z_i, v_i)\}_{i=1}^N$. Here, $(x_i, y_i, z_i)$ are the spatial coordinates and $v_i$ denotes the scalar speed at the $i$-th waypoint.

Our final goal is to formulate a function, $f_{mas}$, that maps a natural language input $\mathcal{L}_{in} = \{\mathcal{L}_1, \dots, \mathcal{L}_K\}$, with $K {\in} \mathbb{R}_+ $, to a modified trajectory $\mathbf{\xi}_{mod}$, which satisfies all user instructions while remaining safe and feasible within the environment. This function must robustly handle the complexity arising from multiple instructions, such as simultaneously satisfying ``move closer to the red bowl'' and ``but stay more above the table''. The inherent challenge of satisfying multiple, interdependent constraints is the primary motivation for our multi-agent system, which explores diverse strategies to find an optimal solution. This process is formulated as 
\begin{equation}
\mathbf{\xi}_{mod} = f_{mas}(\mathbf{\xi}_o, \mathcal{L}_{in}, \mathcal{O}) .
\end{equation}

\subsection{VLM based Geometry Registration}
To acquire geometric information from the scene, we constructed an automated pipeline based on VLM herein, using GPT-4o and foundation vision models. To obtain more comprehensive geometric features and enhance robustness against occlusions, the pipeline utilizes multiple eye-to-hand RGB-D cameras. These cameras are stereo-calibrated to enable point cloud fusion, with the accuracy further refined using the Iterative Closest Point (ICP) algorithm. 
The pipeline then leverages the visual reasoning capabilities of GPT-4o to capture image to text, and text to the most likely corresponding geometric primitive given the existing environment and contexts.

For each identified object, we are using Grounding-DINO\cite{liu2024grounding} and Segment-Anything 2 \cite{ravi2024sam} to acquire its mask from different viewpoints. Based on these masks, the object's point cloud is segmented from each respective view. After fusing the segmented point clouds, we first perform outlier removal. Following this, the DBSCAN \cite{deng2020dbscan} clustering algorithm is applied to further extract the target object's point cloud and filter out noise. Finally, a 6D Oriented Bounding Box (OBB) is generated from the cleaned point cloud. Based on the 
OBB and the previously identified object shape, a corresponding geometric primitive is generated in the robot's base coordinate system. Our proposed visual registration pipeline includes common shapes such as cubes, cylinders, spheres, and cones. 


\subsection{Geometric Constraints } 
After registering the geometric primitives for each obstacle, the corresponding closest points can be efficiently computed and updated based on the analytic relationships between the robot point-mass and each primitive. More importantly, the formulation provides a semantic and structured representation of the environment, rather than relying on raw point-based data. Each primitive (e.g., sphere, plane, cylinder, cuboid, or conic) inherently encodes interpretable geometric meaning, such as surface normals, axes, and boundaries, which allows the environment to be described and interacted with in a consistent and physically meaningful way.


To ensure reproducibility and improve clarity, we adopt the analytic formulations of closest-point computations as defined in GeoPF~\cite{gong2025geopfinfusinggeometrypotential}. The closest point between each geometric primitive and a robot point mass is analytically determined based on their relative positions in 3D Cartesian space. Specifically, the closest point from a line to a point mass depends on whether the orthogonal projection of the point lies within the line segment or outside its endpoints. A similar projection-based principle applies to the plane case. For the cube, the closest point is obtained by identifying the nearest plane surface among the six faces through repeated plane-to-point evaluations. In the case of a cylinder, the closest point is determined by comparing distances to both the circular end caps and the curved lateral surface.

By modeling obstacles using canonical geometric primitives, the environment gains a richer, structured, and computationally efficient representation. The resulting forces or constraints 
are computed analytically, leading to stable and predictable interactions with the environment. This structured representation underlies the richer obstacle definition, distinguishing our approach from point-
or mesh-based methods 
through 
semantic and geometric consistency 
within the 
environment model.

\begin{algorithm}[t]
\caption{Traj. Refinement via Iterative Force Integration}
\label{alg:trajectory_refinement}
\begin{algorithmic}[1]
\Require 
Initial waypoints $\{\mathbf{w}_i^{(0)}\}_{i=1}^{N}$; \\
Constraint set $\mathcal{C} = \{\mathcal{O}, L_{in}\}$; \\
Coefficients $k$, $k_{\mathrm{ang}}$, $\mathbf{k}_{\mathrm{cos}}$, $k_{\mathrm{self}}$; step size $\eta$; 

\For{$j = 0$ to $J-1$}
    \For{each waypoint $i = 1$ to $N$}
        \State Compute component forces~(\eqref{eq:wp}, \eqref{eq:ang}, \eqref{eq:ext}, \eqref{eq:self})
        \[
        \mathbf{F}_{i,\mathrm{total}}^{(j)} = 
        \mathbf{F}_{i,\mathrm{wp}}^{(j)} +
        \mathbf{F}_{i,\mathrm{ang}}^{(j)} +
        \mathbf{F}_{i,\mathrm{ext}}^{(j)} +
        \mathbf{F}_{i,\mathrm{self}}^{(j)}
        \]
        \State Update waypoint: $\mathbf{\xi}_i^{(j+1)}=\mathbf{\xi}_i^{(j)}+\eta\cdot\mathbf{F}_{i,\mathrm{total}}^{(j)}$
    \EndFor
\EndFor
\State \Return Modified trajectory $\mathbf{\xi}_{mod}$
\end{algorithmic}
\end{algorithm}

\subsection{Constraint Generation and Satisfaction} 
We employ a separate LLM for user intent understanding, which translates natural language commands into corresponding constraints $\mathcal{C}$. These desired features include: the modification type (e.g., positional changes relative to objects, speed), {target}; the direction of the modification; and its intensity. 

For object-based trajectory modification, the LLM aligns the user's instructions with the objects identified through visual registration. In scenarios with multiple instructions, the LLM is also tasked with generating different constraint sequences and additional importance parameters, which are determined based on the significance and fragility of the involved objects.

After obtaining the constraints, we use a geometric-aware vector field-based method to iteratively refine the trajectory as described in Algorithm~\ref{alg:trajectory_refinement}. The refinement process seeks to minimize an implicit objective function by applying a composite force field to each waypoint. This field comprises internal forces that preserve geometric properties like segment length and curvature, an external force that couples the trajectory to constraints generated from user commands and handles obstacle avoidance—wherein obstacle avoidance is achieved by adding a repulsive force at a very small range of influence to each object in the scene—and a regularization force that penalizes deviation from the initial path. The iterative application of these forces guides the trajectory toward a locally optimal configuration.

The position of each waypoint, $\mathbf{w}_{i} \in R^3, i{=}\{0,\dots,N\}$,  is influenced by a total force vector $\mathbf{F}_{i,\mathrm{total}} {\in} R^4$. At any iteration $j$ (line-3 of Alg.~\ref{alg:trajectory_refinement}), the total force on the waypoint is 

\begin{equation}
\mathbf{F}_{i,\mathrm{total}}^{(j)}=\mathbf{F}_{i,\mathrm{wp}}^{(j)}+\mathbf{F}_{i,\mathrm{ang}}^{(j)}+\mathbf{F}_{i,\mathrm{ext}}^{(j)}+\mathbf{F}_{i,\mathrm{self}}^{(j)}.
\end{equation}
%
%
Each component of this force field is described below.

\noindent \textbf {Internal Spring Force ($\mathbf{F}_{i,\mathrm{wp}}^{(j)}$)}: To maintain the intrinsic length of the 
segments, 
an internal spring-like force is introduced.
\begin{align}
\mathbf{f}_{i,\mathbf{wp}}^{(j)} {=} k\left(\|\mathbf{w}_{i+1}^{(j)} {-} \mathbf{w}_{i}^{(j)}\| {-} \|\mathbf{w}_{i+1}^{(0)} {-} \mathbf{w}_{i}^{(0)}\|\right) &
\frac{\mathbf{w}_{i+1}^{(j)} {-} \mathbf{w}_{i}^{(j)}}{\|\mathbf{w}_{i+1}^{(j)} {-} \mathbf{w}_{i}^{(j)}\|},
\nonumber 
\\
\mathbf{F}_{i,\mathrm{wp}}^{(j)}=\mathbf{f}_{i-1,\mathrm{wp}}^{(j)}-\mathbf{f}_{i,\mathrm{wp}}^{(j)}, &
\label{eq:wp}
\end{align}
where $k$ is a scalar stiffness coefficient.

\noindent \textbf {Curvature Regularization Force ($\mathbf{F}_{i,\mathrm{ang}}^{(j)}$):} To ensure smoothness, a force regulates the trajectory's curvature, as follows, 
\begin{align}
& \mathbf{c}_i^{(j)}  
    = \tfrac{1}{2}\left(
            \mathbf{w}_{i+2}^{(j)} -
            \mathbf{w}_{i+1}^{(j)}\right)
    -
    \tfrac{1}{2}\left(
            \mathbf{w}_{i+1}^{(j)} -
            \mathbf{w}_i^{(j)}\right)
\nonumber 
\\
& \mathbf{F}_{i+1,\mathrm{ang}}^{(j)} =k_{\mathrm{ang}}\left(\|\mathbf{c}_i^{(j)}\|-\|\mathbf{c}_i^{(0)}\|\right)\frac{\mathbf{c}_i^{(j)}}{\|\mathbf{c}_i^{(j)}\|},
\label{eq:ang}
\end{align}
where $k_{\mathrm{ang}}$ is the angular stiffness coefficient.

\noindent \textbf {External Constraint Force ($\mathbf{F}_{i,\mathrm{ext}}^{(j)}$):} Derived from one or more cost functions, $\mathcal{G}$, which encode information about object information $\mathcal{O}$, and constraints $\mathcal{C}$ and incorporate geometric information via the closest-point computation between waypoints and objects

\begin{equation}\mathbf{F}_{i,\mathrm{ext}}^{(j)}=\mathbf{k}_{\mathrm{cos}}\cdot\mathcal{G}(\mathbf{w}_{i}^{(j)},\mathcal{O},\mathcal{C}),
\label{eq:ext}
\end{equation}
where $\mathbf{k}_{\mathrm{cos}}$ are the constraint weighting factors determined by importance and intensity.

\noindent \textbf {Self-Adherence Force ($\mathbf{F}_{i,\mathrm{self}}^{(j)}$):} 
A regularization term is included to prevent excessive deviation from the initial path.
\begin{equation}\mathbf{F}_{i,\mathrm{self}}^{(j)}=k_{\mathrm{self}}\left(\mathbf{w}_i^{(j)}-\mathbf{w}_i^{(0)}\right),
\label{eq:self}
\end{equation}
where $k_{\mathrm{self}}$ is the adherence weighting factor. At each iteration $j$, the position of every waypoint is updated based on the total calculated force.
\begin{equation}\mathbf{\xi}_i^{(j+1)}=\mathbf{\xi}_i^{(j)}+\eta\cdot\mathbf{F}_{i,\mathrm{total}}^{(j)},
\end{equation}
where $\eta$ is a learning rate or step-size parameter. This iterative process is repeated for a fixed number of steps to get the modified trajectory.


\subsection{Multi-agent system} 
To address scenarios involving multiple, potentially concurrent instructions of the same type, which may conflict and prevent the target constraints from being achieved, we have designed a multi-agent framework built around four core strategies.
\footnote{These are the base agents of our system, which can be further integrated and extended. For brevity and clarity, herein, we only consider core agents.}  
%
\textbf{(i)} \textit{Parallel Agent}: Treats all target instructions as equal, applying all corresponding geometric-aware vector fields simultaneously;  
\textbf{(ii)} \textit{Sequential Agent}: Executes all instructions sequentially in the order provided by the user, modifying the trajectory iteratively with multiple calls;  \textbf{(iii)} \textit{Sequential Agent with Priority}: A Sequential agent that utilizes a LLM to reorder user instructions based on the importance and fragility of the involved objects (informed by the LLM).  
\textbf{(iv)} \textit{Parallel Agent with Importance}: A parallel agent that receives an additional 'intensity' input for each constraint, with the value determined by the LLM based on importance and fragility.

Our pipeline processes the user's command through a structured, multi-stage system. We first leverage the LLM to decouple 
$\mathcal{L}_{in}$ instructions  into a set of $K$ individual, verifiable constraints, denoted as $\mathcal{C} = \{c_1, c_2, \dots, c_K\}$. Each constraint $c_k$ encapsulates a specific desired modification (e.g., type, target, intensity). This set of constraints forms the basis for trajectory generation, evaluation, and refinement.

\noindent \textbf{Initial Planner:} The Initial Planner receives the set of constraints $\mathcal{C}$ from the LLM. Its responsibility is to translate each individual constraint $c_k \in \mathcal{C}$ into a corresponding geometric-aware vector field. These vector fields are then passed to the agents to guide the trajectory modification process.

\noindent \textbf{Observer:} 
The modified trajectory $\xi_{\text{mod}}$ is evaluated by the Observer against the original one and the set of constraints $\mathcal{C}$. 
For each constraint $c_k$, it computes a satisfaction boolean, 
$S(\xi_{\text{mod}}, c_k) \in \{\text{True, False}\}$. A trajectory is deemed successful only if all constraints are satisfied, i.e., $\forall c_k \in \mathcal{C}, S(\xi_{\text{mod}}, c_k) = \text{True}$. The calculation of $S$ is dependent on the constraint type. For instance, it can be a distance check for proximity constraints, a Cartesian check for directional adjustments, or a speed check for velocity changes.

\noindent \textbf{Refinement Planner:} If the Observer determines that a trajectory fails to satisfy the full set of constraints $\mathcal{C}$, the Refinement Planner initiates an iterative adjustment process. It employs a two-level strategy:
\begin{itemize}
    \item If individual constraints in $\mathcal{C}$ are met while overall fail, the planner adjusts meta-parameters (e.g., intensity, importance) associated with the failing constraints to resolve potential conflicts.
    \item If a specific constraint $c_k$ cannot be satisfied in isolation, the planner infers an issue with its corresponding vector field (e.g., the target is out of default range) and increase the magnitude of the parameters, such as the field's influence range.
\end{itemize}
This refinement loop iterates for a specified number of cycles or until a trajectory that satisfies all constraints in $\mathcal{C}$ is found.

\section{Experiments }
\begin{figure*}[h!]
		\centering
		\includegraphics[width=19cm]{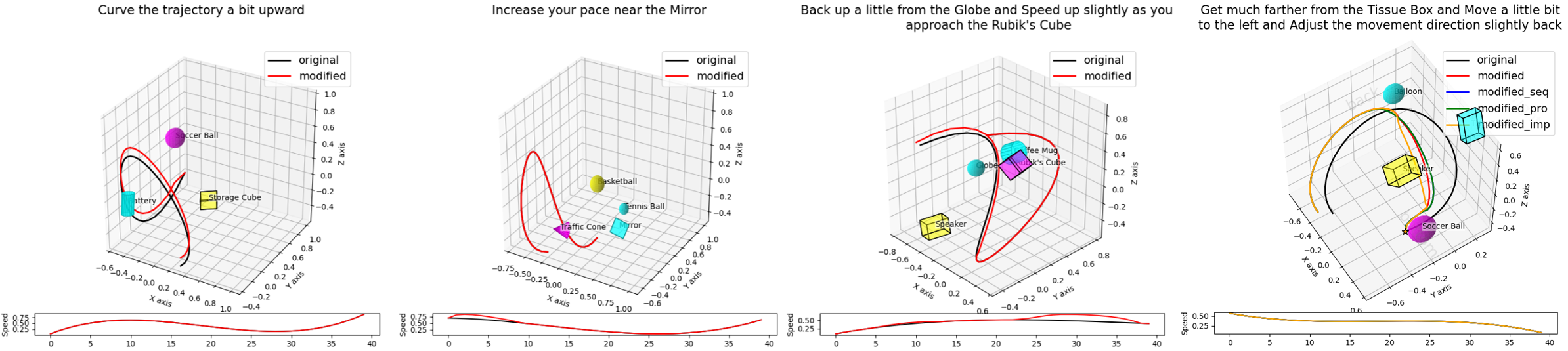}
		\caption{
        %
        Representative results across three synthetic datasets, i.e., single-command, multi-command, and concatenated-instruction inputs. Initial trajectory (black) is modified with \methodname~(red).
        The rightmost image shows the output results of different agents, illustrating how strategy choice (parallel vs. sequential; priority/importance) affects the final reshaped trajectory.} \label{sim1}
\end{figure*}

We conducted extensive experiments in both simulation and real-world settings to validate our framework. Our evaluation had three primary objectives. \textbf{(i)} Quantitatively evaluate the \textit{trajectory adaptation performance} 
across various datasets and task scenarios; \textbf{(ii)} Assess the contribution 
key-components, notably the use of geometric primitives and the multi-agent strategy. Perform both \textit{ablation studies and compare against State-of-the-art (SoTA) baselines};  
and \textbf{(iii)} Validate \methodname's practical \textit{deployment in real-world}, 
and measure user's preferences and perceived performance in terms of safety, intuitiveness, and alignment with user intent. 

\subsection{Dataset} 

To evaluate a wide range of scenarios, we adopt a synthetic trajectory-modification dataset generation procedure (similar to \cite{bucker2022latte,bucker2022reshaping}). 
Each data sample consists of an initial trajectory ($\xi_o$), a language-instruction ($\mathcal{L}_{in}$), and environment objects $O = {O_1, \ldots, O_M}$, each with a 6-DoF pose and an associated geometric shape. 
%
The language instructions are drawn from the three  categories defined in~\cite{bucker2022latte}, i.e.,  
%
(i) absolute Cartesian adjustments (e.g., “go more to the right”), 
(ii) speed modifications (e.g., “slow down when next to the table”), and 
(iii) object-relative repositioning (e.g., “move closer to the chair”).

Using this method,, we curated three distinct datasets variants to stress-test our approach. \textbf{(DS.1)} A \textit{Single-Command Dataset}, analogous to the configuration used in LATTE~\cite{bucker2022latte,bucker2022reshaping}; 
\textbf{(DS.2)} A \textit{Multi-Command Dataset}, extending existing datasets; 
and, finally, 
\textbf{(DS.3)} A \textit{Concatenated-Instruction Dataset}, where each input is a compound instruction containing multiple concatenated commands (testing the model’s ability to handle complex, multi-constraint directives in one query).   
%
 %
 These datasets enable evaluation of \methodname~under progressively more complex language inputs and trajectory constraints.

\subsection{Quantitative Results and Analysis} 

We first examine 
trajectory adaptation in several simulated scenarios across the above datasets. Fig.~\ref{sim1}. illustrates representative outcomes. 
For each, an initial trajectory  (black) is given, and the robot then receives a language command with the desired modification (e.g., moving farther from or closer to an object, or going faster or slower in its vicinity). Our model then yields a revised trajectory (blue) that adheres to the specified spatial and speed profile changes. In the rightmost panel of Fig.~\ref{sim1}, we additionally show outputs from different agents in our multi-agent system, highlighting how each agent pursues the command under varying strategy parameters.
These examples confirm that \methodname~produces feasible trajectory modifications that reflect the instruction's geometric and kinematic intent.

\begin{figure}[t!]
		\centering
		\includegraphics[width=9cm]{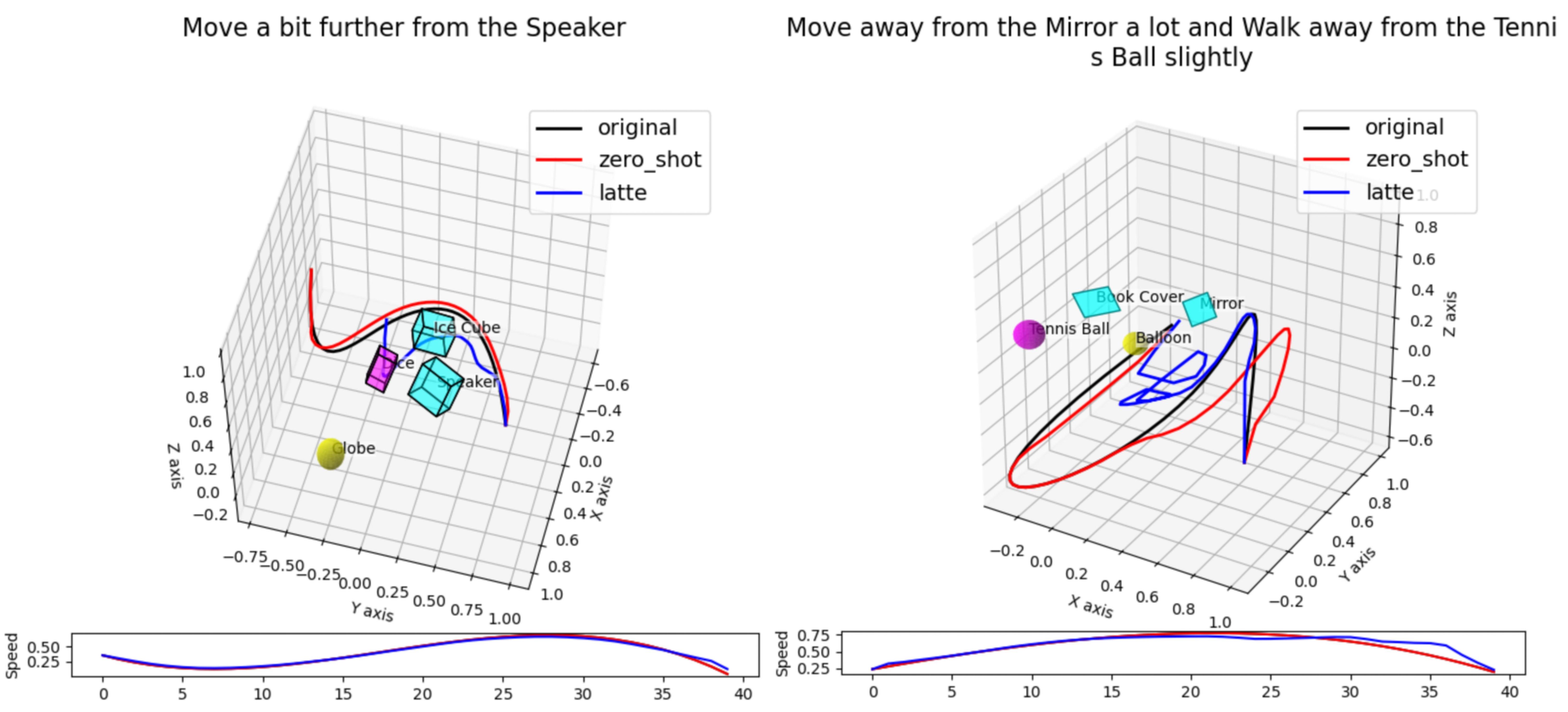}
		\caption{Comparison between LATTE (blue) and \methodname~(red) on the single-command and multiple-command datasets.} 
		\label{compare}
    \end{figure}

\textbf{Comparison to Learning-Based Baseline}: We benchmarked our method against the prior SoTA language-driven approach, LATTE~\cite{bucker2022latte,bucker2022reshaping}. 
Tests were performed on two standard datasets, (a) a single-command (identical to the original LATTE study) and (b) a multi-command set. Since LATTE does not account for object geometry, we provided it with the centroid of our registered geometry as the object's pose. We then applied the same normalization procedure described in~\cite{bucker2022latte,bucker2022reshaping} to ensure output consistency and fairness.

As shown in Fig.~\ref{compare}, our learning-free, zero-shot geometric-aware strategy produces smoother and more stable trajectory modifications than the deep learned LATTE policy. 
By explicitly encoding geometric and kinematic constraints, our method guarantees that the user’s instructions are satisfied by the optimized trajectory. This advantage is especially pronounced in multi-command scenarios: \methodname~handles successive instructions without conflating their effects, maintaining a predictable modification for each part of the command. In contrast, the LATTE struggles with open-vocabulary and multi-commands---often yielding erratic or suboptimal paths. 



\begin{figure*}[h!]
		\centering
		\includegraphics[width=19cm]{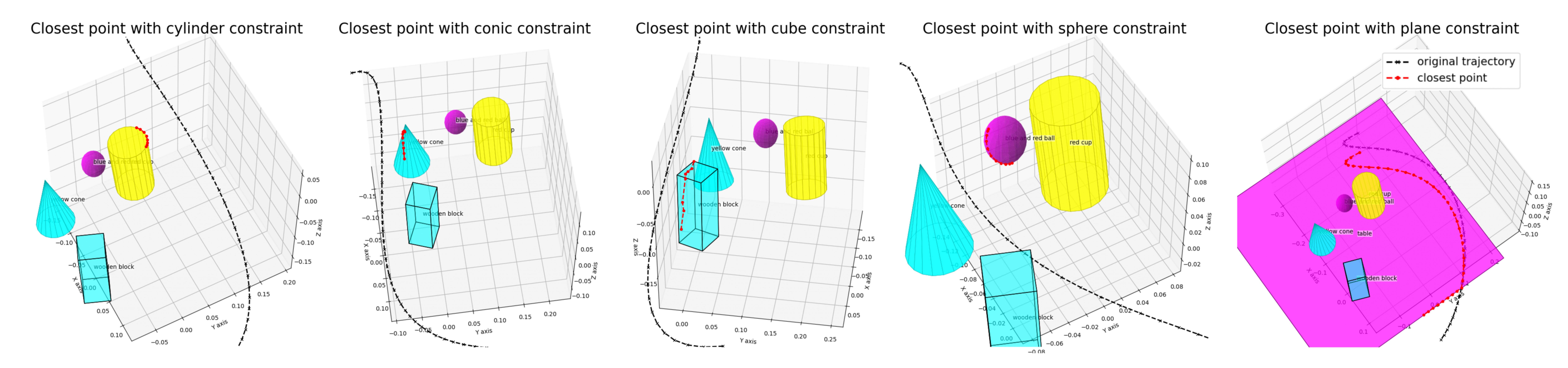}
		\caption{The image shows the points along the trajectory and their corresponding nearest points on the object. When the distance exceeds a certain threshold, no nearest point is assigned. Black indicates the trajectory, and red indicates the nearest points.} \label{closest}
\end{figure*}

\textbf{Effect of Geometric Object Representation}: A key feature of our model is its use of full object geometry (via primitive shapes). 
Fig.~\ref{closest} shows examples where a series of closest-point markers (red) lie along objects surface next to the trajectory (black) at various waypoints. This, together with the primitive forces, yields smoother and nuanced interactions along the path.  
This contrasts with SoTA alternatives, which depict an object only by a single reference point (e.g., center of its 3D bounding box). 
 These are offset from the object’s surface and can mislead the language-driven optimizer (especially for end-to-end learning) by misjudging distances. 


\begin{figure}[t!]
		\centering
		\includegraphics[width=9cm]{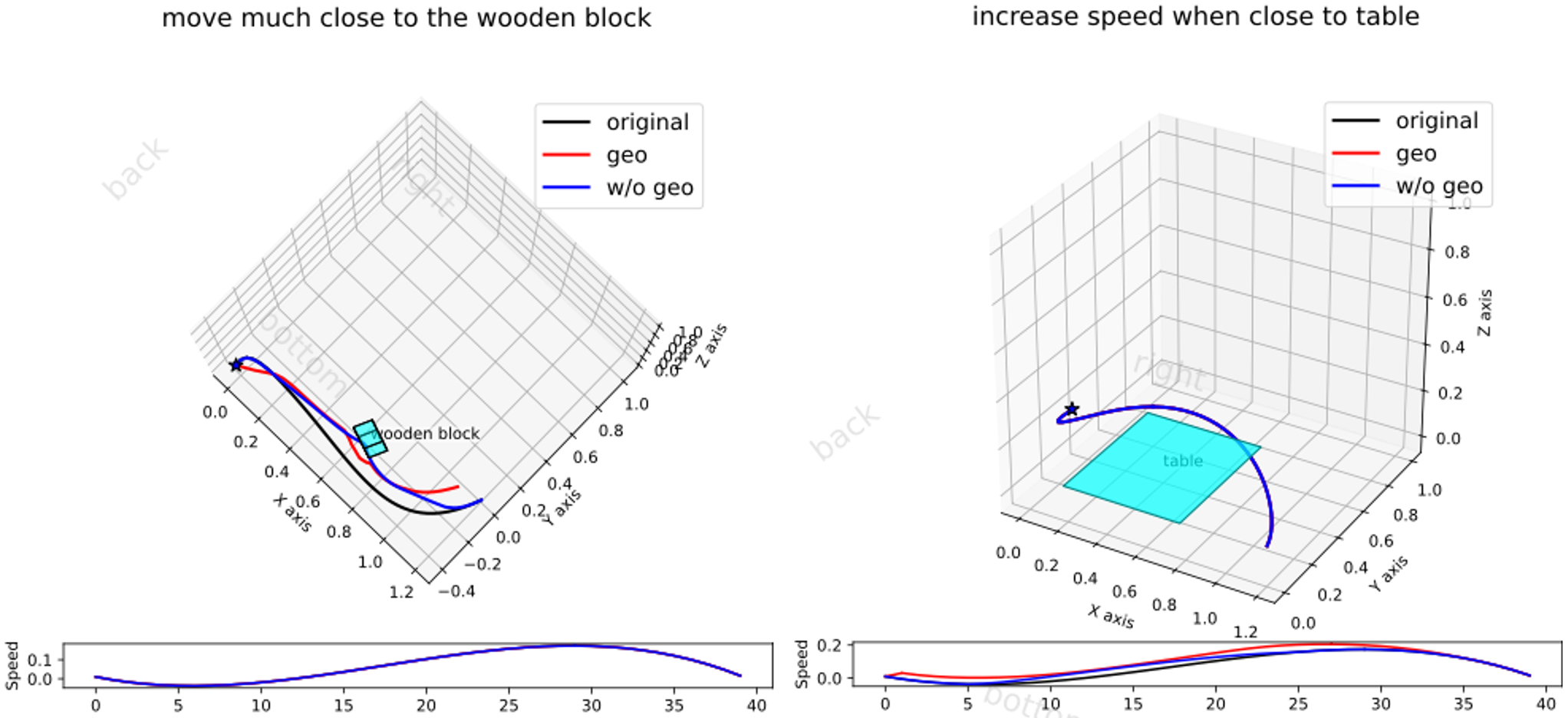}
		\caption{Comparison between geometry-based trajectory adaptation and non-geometry-based adaptation.} 
		\label{geo}
    \end{figure}

Fig.~\ref{geo} compares our method (red) against a point-based method (blue).   
For fairness, both were driven by the same LLM-generated instructions.  
In the left, “move closer to the object” leads the point-based method to naively generate a trajectory that penetrates the object. 
%
%
In stark contrast, our geometry-aware method computes the nearest surface points efficiently via the primitives, and  keeps the trajectory at a safe distance. 
Similarly, in Fig.~\ref{geo}(right), 
SoTA methods only adjust velocity near the table’s center, while our  
%
approach 
correctly applies the speed change throughout the entire region of influence defined by the table's geometry. These results underscore the critical role of the geometric information for accurate, safe, and human-aligned trajectory modification.

\begin{figure}[t!]
		\centering
		 \includegraphics[width=1\columnwidth]{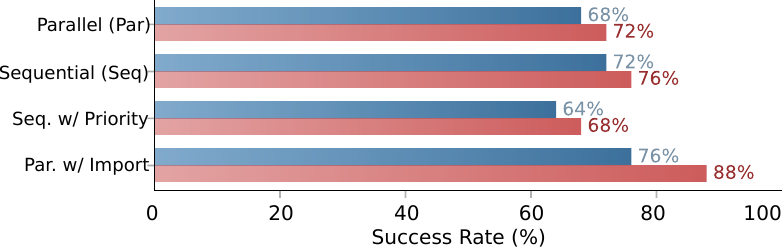}
        \caption{The success rate of different agents with a single iteration (\textcolor{blue}{\textbf{---}}) and after the final round of feedback-refinement (\textcolor{red}{\textbf{---}}).}
		\label{agent1}
        \vspace{-10pt}
    \end{figure}

\begin{figure}[t!]
		\centering
            \includegraphics[width=1\columnwidth]{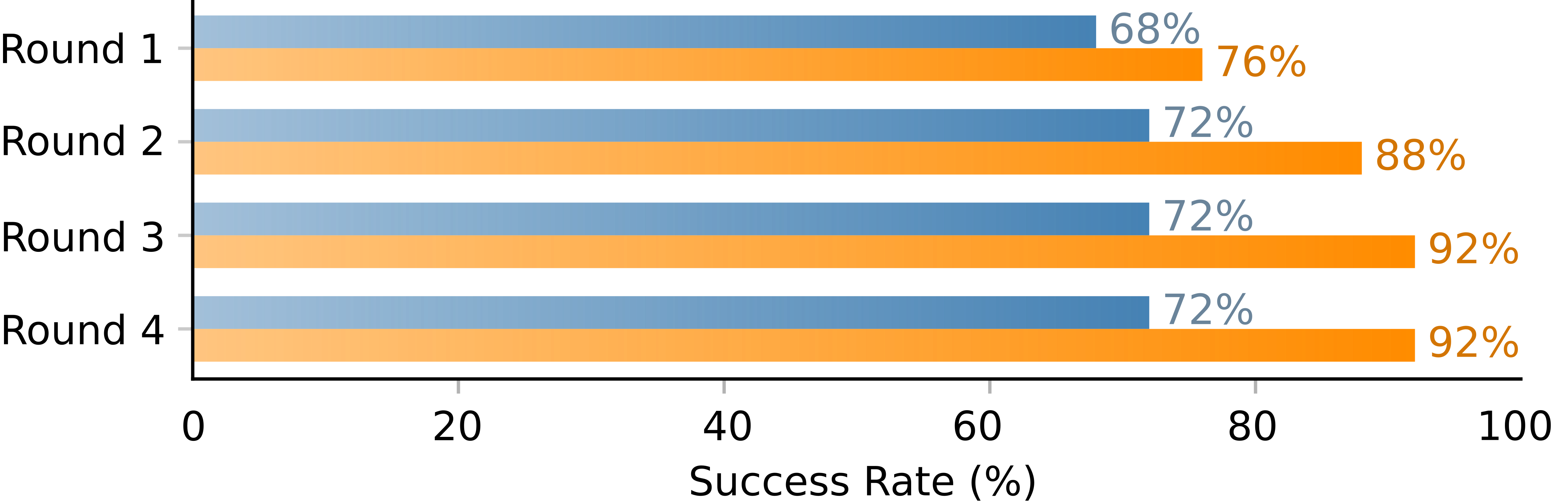}
		\caption{The success rate of the single agent (\textcolor{blue}{\textbf{---}}) and multi-agent systems (\textcolor{orange}{\textbf{---}}) after multiple refinement rounds.} 
		\label{agent2}
    \end{figure}

\textbf{Effect of the Multi-Agent Strategy}: To evaluate \methodname’s multi-agent refinement mechanism, we devised 
%
challenging trajectory modification tasks using datasets (b)-(c) as described above. In our setting (Fig.~\ref{fig:model_arc}, bottom-right), 
the LLM provides an initial set of different constraints from the language command, and an observer module checks whether a given trajectory meets them. 
The observer verifies satisfaction using three quantitative checks:   
(i) a \textit{distance check} that compares the average distance between the trajectory and the target object before vs after modification (using five closest waypoints); (ii) a Cartesian check that verifies the displacement along a specified direction; and (iii) a speed check that examines the change i velocity (same 5 waypoints).  An agent is deemed valid only if all relevant checks pass. 
Multiple agents, with different prioritization or sequencing policy, run in parallel. If no agent succeeds, refinement cycles are triggered to adjust parameters and iterate once again with the multi-agent system. 
Herein, our iteration is limited to three rounds 
as we observed a performance plateau---mostly due to intrinsically conflicting constraints in such complex tasks. 
As shown in Fig.~\ref{agent1}, all agents improve their success rates after refinement. Fig.~\ref{agent2} confirms that the multi-agent setup consistently outperforms both single-agent runs and even the best-performing individual agent.  These results validate the design’s efficacy for resolving ambiguity, prioritizing competing goals, and improving success in long-horizon, multi-instruction scenarios


\begin{figure}[t!]
		\centering
            \includegraphics[width=8cm]{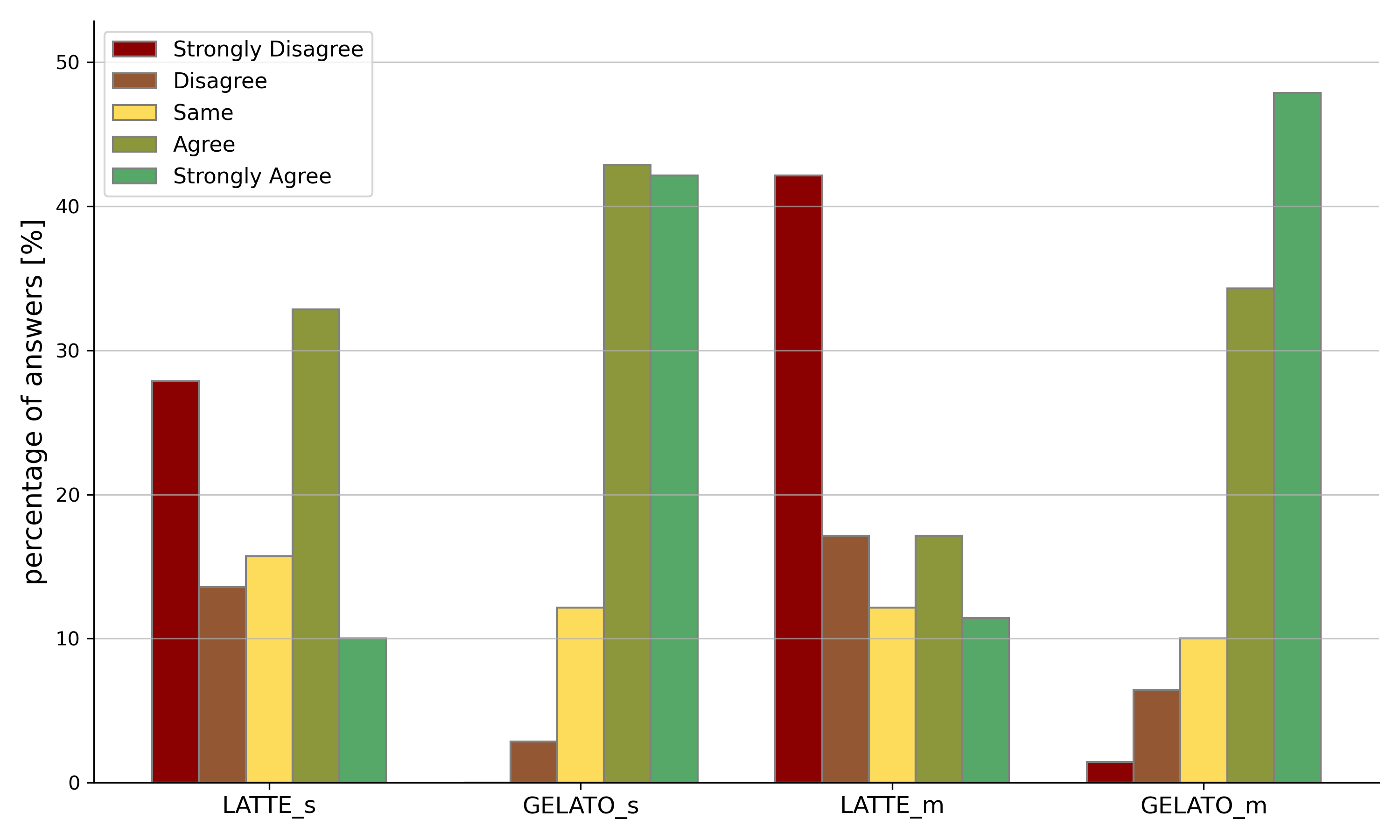}
        \caption{Distribution of 400 Likert-scale ratings (10 participants) comparing trajectory modifications across methods under single- and multi-instruction prompts.}
		\label{user_study}
    \end{figure}

\subsection{User study experiments} 
Beyond objective metrics, we assessed 
performance via a user study.  
We recruited 10 volunteer participants (non-experts) and collected a total of 400 data point ratings.  
%
Each participant rated, on a 1–5 Likert scale, the trajectory modifications produced by different approaches for a given natural language interaction, including prior SoTA methods under both the single-command and multi-command settings. Fig.~\ref{user_study} illustrates the distribution of responses across the baselines. The results indicate a clear preference for our approach, as \methodname~achieved the highest scores on average in both single and multi-instruction settings. Users consistently found \methodname’s trajectory adjustments to be more intuitive and closer to the intended behaviour. In contrast, the original LATTE received notably lower ratings, particularly in the open-vocabulary multi-command tasks---reflecting the technical results from the quantitative assessment. 
These results reinforce that \methodname’s geometry-aware, constraint-driven method yields not only objectively safer trajectories but also motions that humans perceive as more appropriate and aligned with their requests.




\subsection{Real-World Experiments} 


Experiments were conducted on a 7-DOF Franka Robot equipped with a parallel gripper. The control system ran on an Intel i9-14900K CPU and NVIDIA RTX 4060 GPU. Dual Intel RealSense D435 RGB-D cameras captured stereo views of the workspace, providing dense point clouds. For semantic grounding and instruction interpretation, we used a GPT-4-based vision-language model via API, along with either GPT-4 (or the lighter DeepSeek-V3, when indicated) as the LLM. Object segmentation was performed locally using Grounding DINO and Segment Anything 2(SAM2).



In each trial, randomly placed household objects were registered into 3D geometric primitives via \methodname’s automated vision module, Fig~\ref{geo_spot}.  It first segments the scene using the vision models and then performs point-cloud registration to fit primitive shapes to the objects. To ensure consistency, we set a strict threshold of 0.02 m for the ICP alignment algorithm’s convergence, and we only consider segmented object masks with a confidence score above 99\%. 
The raw point clouds are further refined by removing outliers  (20-neighbor statistical threshold, 1.0 standard deviation cutoff), and DBSCAN clustering (radius 0.15 m, minimum 15 points). Each cluster was fitted to a primitive shape (e.g., plane, cylinder, cuboid) based on a GPT-inferred primitive-based category, and mapped to the robot's base coordinate frame.

\begin{figure}[t!]
	\centering
    \includegraphics[width=\columnwidth]{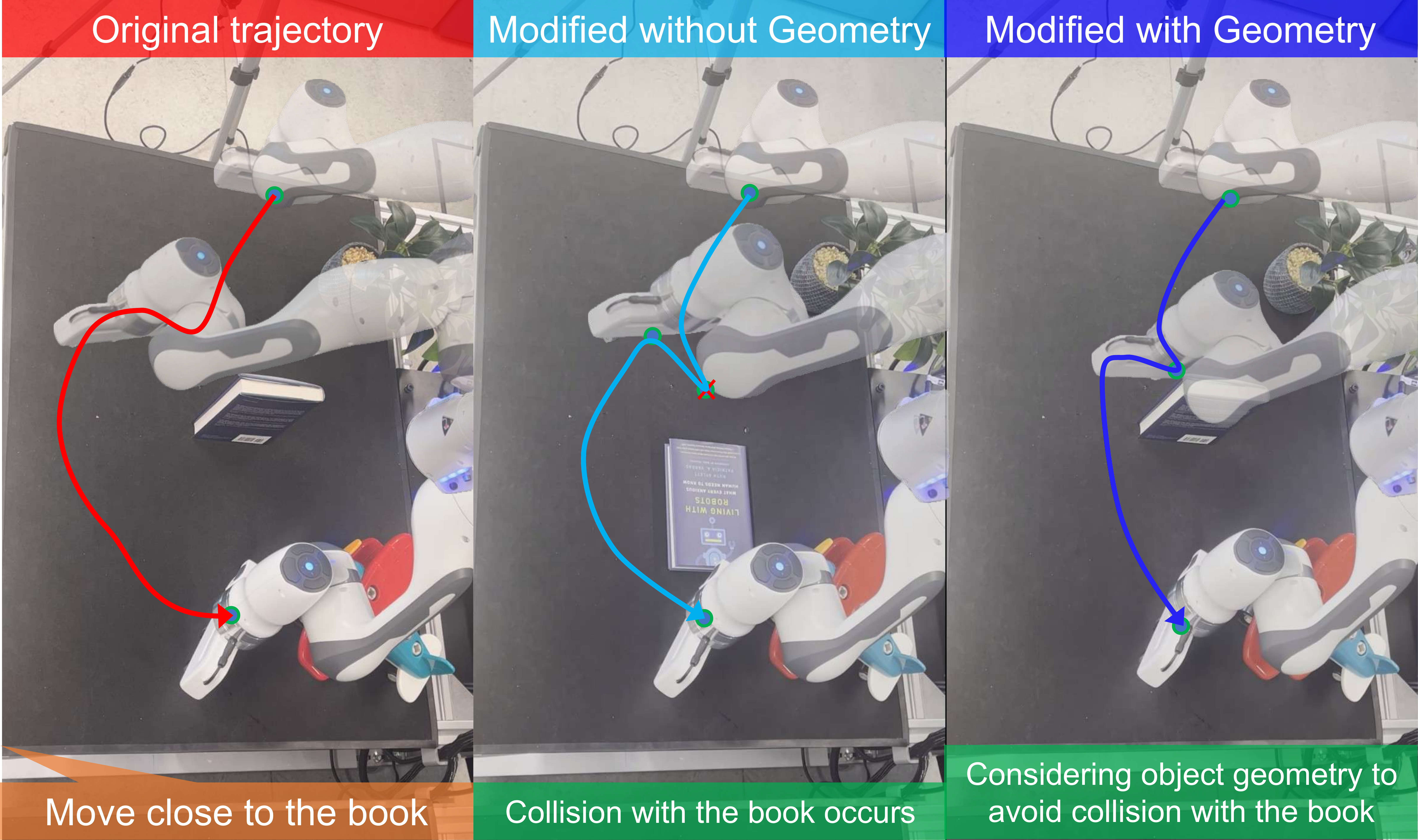} \\[4pt]  
    \includegraphics[width=\columnwidth]{figures/geo_table.pdf} \\[4pt]    
    \includegraphics[width=\columnwidth]{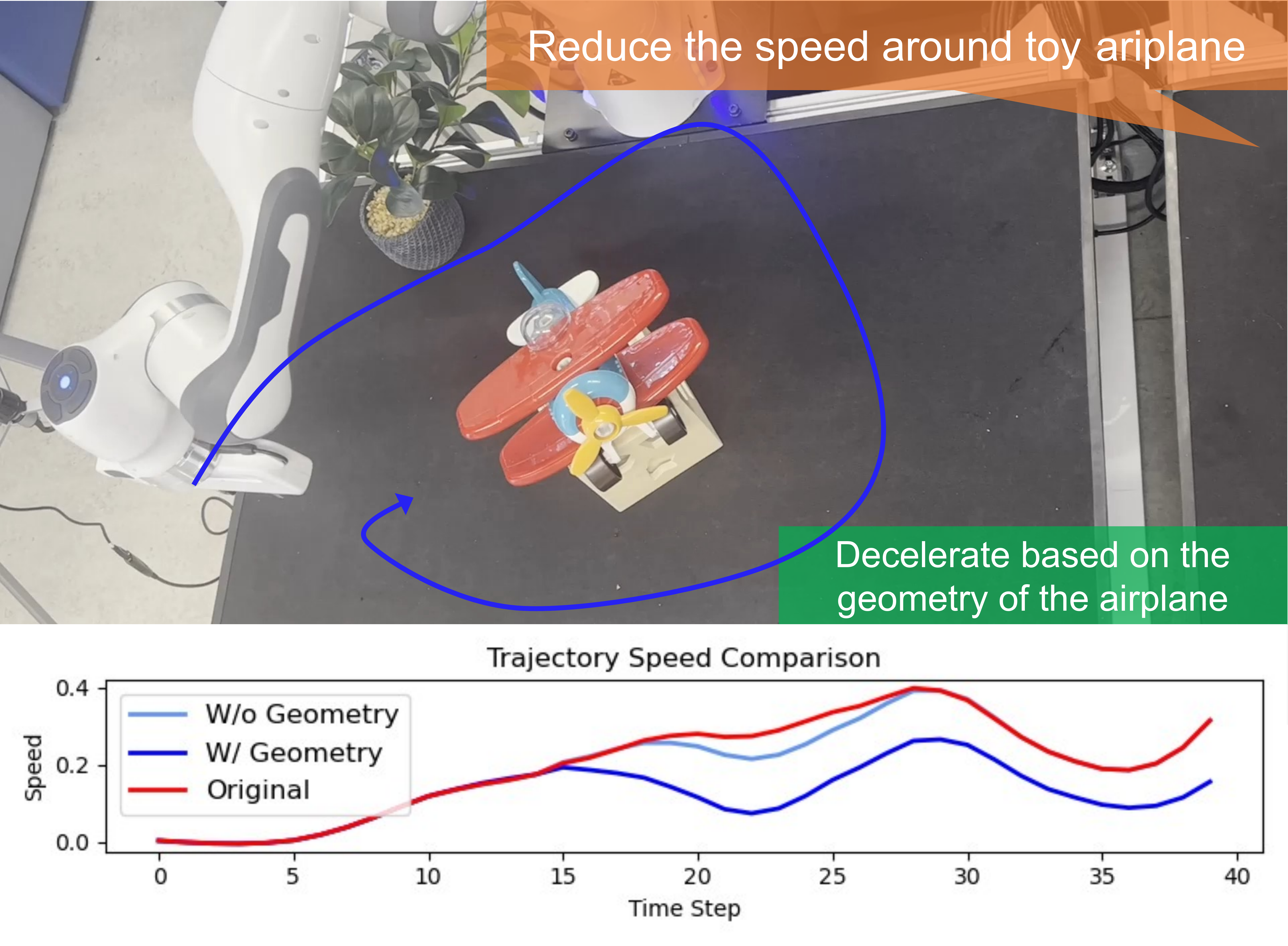}
    \caption{
Three real-world scenarios. \textbf{Top}: Highlights point-proxy language-driven methods leads to collision, while our approach secures a safety-distance. \textbf{Middle}: 
“Get closer to the table” command with the baseline leads to changes only near the table centroid, while our method uniformly reshapes height across the entire tabletop. \textbf{Bottom}: Similarly, existing language-driven methods only reduce the speed at a single point, while \methodname~ensures the desired behaviour  throughout the airplane’s geometry. 
    }
	\label{compare_task}
\end{figure}


\textbf{Results:}  
Our real-world experiments reinforce the results from our quantitative assessment and user studies, and further demonstrate that the geometry-aware trajectory modifications offered by \methodname~are not only feasible on actual hardware but also crucial for safety, performance, and adherence to language commands. 
Fig.~\ref{geo_spot} and \ref{compare_task} highlight 
several comparisons between our method and a geometry-agnostic baseline (one that treats objects as points). 
In the top scenario of Fig.~\ref{compare_task}, the robot is instructed to “move close to the book”. A point-based method, unaware of its geometry, causes a collision with the book. In contrast, 
our automatic geometric-awareness pipeline, accounting for the object’s dimensions in an efficient and explainable manner, successfully gets closer to the book while avoiding any collision.
This underscores how our approach yields safer motions in cluttered spaces. 

In the second scenario (middle of Fig.~\ref{compare_task}), the robot is asked to lower a tall cup toward a tabletop. The baseline only lowers the trajectory at the table’s center, misinterpreting the full geometric semantics of the table (essentially focusing on the table’s centroid). In contrast, \methodname~adjusts the entire trajectory consistently across the table’s surface.
In the third task (bottom), the instruction is to reduce speed around a toy airplane. Again, the point-based method applies the change narrowly at the centroid, but \methodname~modulates speed across the full geometry-aware influence zone.

\section{Conclusion}

We introduced \methodname, a learning-free geometric-aware multi-agent framework that enables robots to reshape their trajectories in response to complex multi-input natural language commands. Our approach leverages VLM for automated geometric registration and utilizes LLM to translate user intent into actionable geometric-aware vector field forces with interactive reasoning and observer-guided refinement. We conducted a comprehensive set of experiments, demonstrating our method's superior performance in safety and accuracy against traditional point-based approaches and its enhanced stability compared to state-of-the-art learning-based models. Additionally, real-world and simulated results validate the necessity of our multi-agent system for robustly handling multi-input / multi-constraint tasks and underscore the critical role of geometry awareness in achieving safe and intuitive human-robot interaction.
Future work targets preference-aware sequencing, tackling potential mismatches from LLM-based generation and user inputs, and uncertainty-aware composite primitive models. 





\printbibliography

\end{document}